\documentclass[sigconf]{acmart}

\usepackage{multirow}
\usepackage{hyperref}

\AtBeginDocument{%
  \providecommand\BibTeX{{%
    \normalfont B\kern-0.5em{\scshape i\kern-0.25em b}\kern-0.8em\TeX}}}

\copyrightyear{2022}
\acmYear{2022}
\setcopyright{acmcopyright}

\acmConference[MM '22]{Proceedings of the 30th ACM International Conference on Multimedia}{October 10--14, 2022}{Lisboa, Portugal}
\acmBooktitle{Proceedings of the 30th ACM International Conference on Multimedia (MM '22), Oct. 10--14, 2022, Lisboa, Portugal}
\acmPrice{15.00}
\acmDOI{10.1145/3503161.3548412}
\acmISBN{978-1-4503-9203-7/22/10}

\acmSubmissionID{mmfp3100}

\begin{document}

\title{Hierarchical Few-Shot Object Detection: Problem, Benchmark and Method}

%

\author{Lu Zhang}
\email{l\_zhang19@fudan.edu.cn}
\orcid{0000-0001-9532-5219}
\affiliation{%
	\institution{Fudan University}
	\city{Shanghai}
	\country{China}
}

\author{Yang Wang}
\email{tongji\_wangyang@tongji.edu.cn}
\affiliation{%
	\institution{Tongji University}
	\city{Shanghai}
	\country{China}
}

\author{Jiaogen Zhou}
\authornote{Correspondence authors: Jiaogen Zhou (School of Urban and Environmental Sciences, Huaiyin Normal University), Shuigeng Zhou (School of Computer Science, and Shanghai Key Lab of Intelligent Information Processing, Fudan University.)}
\email{zhoujg@hytc.edu.cn}
\affiliation{%
	\institution{Huaiyin Normal University}
	\city{Huaian}
	\country{China}
}

\author{Chenbo Zhang}
\email{cbzhang21@m.fudan.edu.cn}
\affiliation{%
	\institution{Fudan University}
	\city{Shanghai}
	\country{China}
}

\author{Yinglu Zhang}
\email{yingluzhang21@m.fudan.edu.cn}
\affiliation{%
	\institution{Fudan University}
	\city{Shanghai}
	\country{China}
}

\author{Jihong Guan}
\email{jhguan@tongji.edu.cn}
\affiliation{%
	\institution{Tongji University}
	\city{Shanghai}
	\country{China}
}

\author{Yatao Bian}
\email{yatao.bian@gmail.com}
\affiliation{%
	\institution{Tencent AI Lab}
	\city{Shenzhen}
	\country{China}
}

\author{Shuigeng Zhou}
\authornotemark[1]
\email{sgzhou@fudan.edu.cn}
\affiliation{%
	\institution{Fudan University}
	\city{Shanghai}
	\country{China}
}

\renewcommand{\shortauthors}{Lu Zhang et al.}

\begin{abstract}
Few-shot object detection (FSOD) is to detect objects with a few examples. However, existing FSOD methods do not consider hierarchical fine-grained category structures of objects that exist widely in real life. For example, animals are taxonomically classified into orders, families, genera and species etc. In this paper, we propose and solve a new problem called hierarchical few-shot object detection (Hi-FSOD), which aims to detect objects with hierarchical categories in the FSOD paradigm. To this end, on the one hand, we build the first large-scale and high-quality Hi-FSOD benchmark dataset HiFSOD-Bird, which contains 176,350 wild-bird images falling to 1,432 categories. All the categories are organized into a 4-level taxonomy, consisting of 32 orders, 132 families, 572 genera and 1,432 species. On the other hand, we propose the first Hi-FSOD method HiCLPL, where a hierarchical contrastive learning approach is developed to constrain the feature space so that the feature distribution of objects is consistent with the hierarchical taxonomy and the model's generalization power is strengthened. Meanwhile, a probabilistic loss is designed to enable the child nodes to correct the classification errors of their parent nodes in the taxonomy. Extensive experiments on the benchmark dataset HiFSOD-Bird show that our method HiCLPL outperforms the existing FSOD methods.
	
\end{abstract}

\ccsdesc[500]{Computing methodologies~Object detection}


\keywords{Few-shot object detection; hierarchical few-shot object detection; Benchmark; hierarchical classification.}

\maketitle

\begin{figure}[t]
	\centering
	\includegraphics[width=0.9\linewidth]{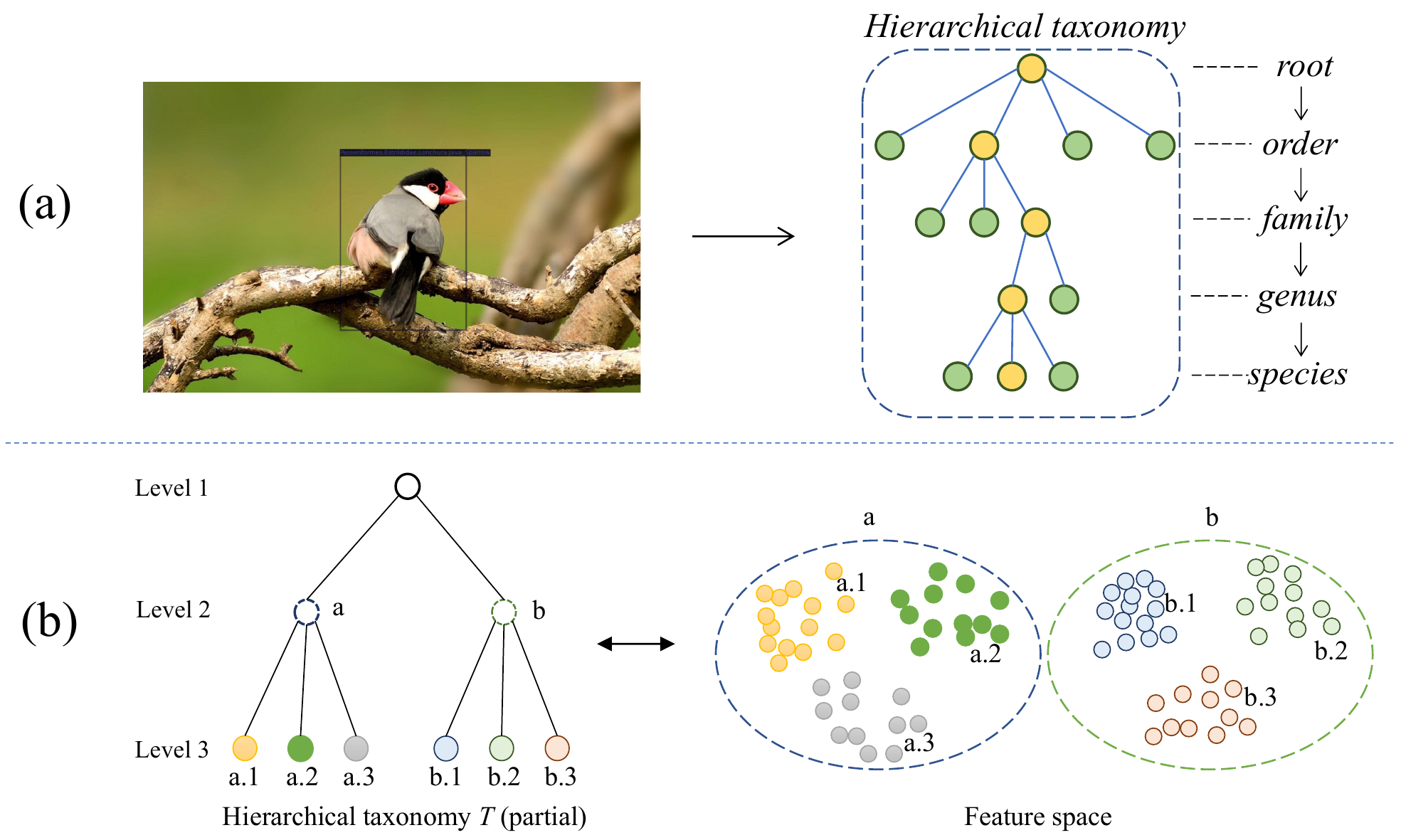}
	\caption{
		(a) The hierarchical taxonomy of our HiFSOD-Bird dataset;
		(b) Illustration of the proposed hierarchical contrastive learning, which constrains the feature space such that the distribution of object features is consistent with the hierarchical taxonomy. 
	}
	\label{fig:intro}
\end{figure}

\section{Introduction}
Existing object detection methods~\cite{FasterRCNN-NIPS} require huge amounts of annotated training data. However,
in the real world, samples of some categories are difficult to acquire and the cost to label high-quality samples can be very high.
On the contrary, a child can recognize and locate elephants or horses in a picture that s/he has never seen before with only a few examples.
Thus, few-shot object detection (FSOD)~\cite{TFA,FSCE,Attention-RPN,DeFRCN,MPSR,Relation_Reasoning,Meta-DETR,MetaRCNN,liutianying,Han_2022_CVPR} is gaining increasing research interests, which tries to detect novel
objects with only a few labeled examples.
However, many objects in real life fall into hierarchical fine-grained category structures. For example, elephants have different families and species, e.g. African elephants and Asian elephants. And African elephants have two subspecies, which are African savannah elephants and African forest elephants, so as the Asian elephants. Obviously,
it is difficult for an ordinary people (let alone a child) to distinguish between an African savannah elephant and an African forest elephant if only a few photos are given.
Moreover, existing FSOD methods do not consider such hierarchical fine-grained category structures of objects that exist ubiquitously in real life, thus they cannot cope with such scenarios well.
In this paper, we propose a new problem of \textbf{hi}erarchical \textbf{f}ew-\textbf{s}hot \textbf{o}bject \textbf{d}etection, \textbf{Hi-FSOD} in short, which aims to perform few-shot object detection under a hierarchical taxonomy. Obviously, the FSOD task is a special case of Hi-FSOD when the hierarchical taxonomy is degenerated to a flat category structure. So comparing to FSOD, Hi-FSOD is more challenging and has wider applications than FSOD, especially in the scenarios that the number of categories of objects is huge, where existing FSOD methods are neither efficient nor effective.
To address the Hi-FSOD problem, we have tackled two major subproblems:

On the one hand, we construct the first
high-quality and large-scale Hi-FSOD benchmark dataset of wild birds, which is called \textbf{HiFSOD-Bird}.
Although there are already some datasets of wildlife for computer vision (CV) tasks~\cite{CUB,AwA,AP-10k,FishDataset}, most of them are for classification tasks and a few of them are dedicated to object detection tasks.
Nevertheless, few of them have a strictly hierarchical organization of categories.
Existing FSOD methods perform training and testing on the modified COCO~\cite{COCO} and VOC~\cite{VOC} datasets whose label structures are flat and contain only 80 and 20 categories, respectively, which thus are unsuitable for the Hi-FSOD task.
Our HiFSOD-Bird dataset contains totally 1,432 categories and 176,350 bird images with high-quality annotated bounding boxes. All categories are organized into a 4-level hierarchical taxonomy: from top to bottom, order, family, genus and species,
%
as shown in Fig.~\ref{fig:intro}(a).
It consists of 32 orders, 132 families, 572 genera and 1,432 species, covering more than 90\% of the world's water birds and part of forest birds.
The bounding boxes and class labels of each image are manually annotated and carefully double-checked.
Moreover, each category of birds comes with a textual description, so the dataset can be further used for the zero-shot object detection task.
The HiFSOD-Bird dataset is also of great significance to the monitoring and protection of endangered birds, since the samples of endangered birds are difficult to acquire and the domain knowledge is mainly from expert annotations.

On the other hand, we develop the first Hi-FSOD method \textbf{HiCLPL}, which is a two-stage method with \textbf{hi}erarchical \textbf{c}ontrastive \textbf{l}earning and \textbf{p}robabilistic \textbf{l}oss. Here, hierarchical contrastive learning (HiCL) is used to constrain the feature space so that the feature distribution of objects is consistent with the hierarchical category structure,  and the probabilistic loss is designed to enable the child nodes to correct the classification errors of their parent nodes.
Fig.~\ref{fig:intro}(b) illustrates the HiCL mechanism. We use memories to hold the prototypes of classes in the hierarchical tree. Then, a hierarchical contrastive loss is designed to control the distance between box features and memories at different levels. Finally, we utilize exponential moving average to update the parameters of memories. HiCL can boost the generalization power of the model.
Meanwhile, we found that in the process of hierarchical classification from top to bottom, if a non-leaf node wrongly classifies an instance, the classifications of the instance at the descendants nodes are useless. Therefore, we design a probabilistic loss such that the child nodes can learn to identify and correct the misclassified samples of their parent nodes.

In summary, contributions of this paper are as follows:
1) We propose a new problem of hierarchical few-shot object detection (Hi-FSOD), which is an extension to the existing FSOD problem, so it is more challenging and has wider applications.
2) We establish the first large-scale and high-quality benchmark dataset HiFSOD-Bird,  specifically for the Hi-FSOD problem.
3) We develop the first Hi-FSOD method HiCLPL, which uses hierarchical contrastive learning to constrain the feature space and a probabilistic loss to correct the classification errors of parent nodes.
4) We conduct extensive experiments on the benchmark dataset HiFSOD-Bird to evaluate the proposed method HiCLPL. Experimental results show that our method HiCLPL outperforms the existing FSOD methods.

\section{Related Work}\label{sec:related_work}
\subsection{Few-shot Object Detection}
Existing few-shot object detection (FSOD) methods roughly fall into two types:  meta-learning based and fine-tuning based.
Meta-learning based methods~\cite{Feature-Reweighting,MetaRCNN,Attention-RPN,Meta-DETR, SQGuidance} learn meta knowledge from base classes to facilitate model training for novel classes. Among them,
FSRW~\cite{Feature-Reweighting} utilizes
a feature re-weighting strategy to construct a one-stage object detector.
Attention-RPN~\cite{Attention-RPN} integrates the information of supports into RPN, in order to pay more attention to the foreground objects relevant to support classes. Meta-DETR~\cite{Meta-DETR} exploits the inter-class correlation to apply the detection transformer~\cite{deformableDETR} to the FSOD task. We proposed a support-query mutual guidance strategy that can generate more support-relevant candidate regions, together with a hybrid loss to enhance the metric space~\cite{SQGuidance}.
Fine-tuning based methods~\cite{TFA,MPSR,Relation_Reasoning,DeFRCN,FSCE} formulate the FSOD problem in a transfer learning setting. TFA~\cite{TFA} is the first work that proposes a two-stage fine-tuning strategy. It first trains the entire model on the base classes, and then fine-tunes the final classifier on a balanced dataset containing base and novel data. Experiments show that such fine-tuning method is simple yet very effective. Following TFA, a number of methods are developed. DeFRCN~\cite{DeFRCN} adopts multi-stage and multi-task decoupling to improve performance. FSCE~\cite{FSCE} uses a contrastive learning strategy to constrain the intra-class similarity and enhance the inter-class similarity of box features.
Nevertheless, existing methods do not consider the scenarios where object classes form a hierarchical taxonomy, thus they cannot be directly used to effectively handle the problem proposed in this paper.

Different from these works above, here we address a new problem --- hierarchical few-shot object detection (Hi-FSOD). To this end, we build a large-scale and high-quality benchmark dataset and develop an effective method.

\subsection{Fine-grained Image Recognition}
Fine-grained image recognition (FGIR) is to recognize numerous visually similar subcategories under the same basic category.
Existing FGIR methods can be divided into the following three types: (1) discriminative region based methods
~\cite{Part-Stacked_CNN,Object-Part-Attention,Mask-CNN,Look-Closer,Multi-Attention-Convolutional,Multi-Class-Constraint} adopt a two-stage paradigm that first locates the key object parts and then does  classification based on the discriminative regions, (2)
attention based methods
~\cite{Look-Closer,Multi-Attention-Convolutional,Multi-Class-Constraint} leverage attention mechanisms to automatically localize discriminative regions of fine-grained objects, and (3)
loss function based methods ~\cite{Maximum-Entropy-Fine,Channel-Interaction,Subtle-Differences} focus on constructing effective loss functions to directly regularize the learnt representations.

Like the FSOD works, existing FGIR methods also perform classification under a flat category structure, but here the difference between categories is visually too slight to be used to easily distinguish the objects of different categories.

\subsection{Hierarchical Classification}
Hierarchical classification (HC) is a challenging problem where the classes are organized into a predefined hierarchy.
HC methods in traditional machine learning mostly follow the top-down strategy that decomposes the hierarchical classification task into a series of subtasks and then trains an independent classifier for each subtask, i.e., transforming a coarse-grained category into several fine-grained subcategories~\cite{taxonomies-visual,Embedding-Trees,Hierarchical-Category-Structure,FastandBalanced}.
For deep hierarchical classification, some works try to directly embed the prior semantic knowledge contained in the class structure to visual features to guide the classification. Among them, \cite{DeViSE} develops a visual-semantic embedding model, which transfers the learnt semantic information to visual object recognition. \cite{Hierarchy-Retrieval} proposes to map images to class embeddings to learn semantically discriminative features. Besides, some approaches introduce the multi-task framework to solve deep hierarchical classification efficiently~\cite{MakingBetterMistakes,Semantics-aware}.
However, hierarchical classification for object detection has received little attention, though hierarchical categories exist ubiquitously in real life.


\section{Problem Definition}\label{sec:problem}
Hierarchical few-shot object detection~(Hi-FSOD) is an extension to few-shot object detection, where the categories are organized hierarchically.
Formally, the class space $C$ is divided into the base classes $C_b$ and novel classes $C_n$, where $C = C_b \cup C_n $ and $ C_b \cap C_n$ = $\varnothing$.
In the base classes $C_b$, each class has many instances while only $K$ (usually less than 10) instances available per category in the novel classes $C_n$.
In our hierarchical setting, each class $c_i \in C$ can be mapped to a leaf node in the hierarchical tree (taxonomy) $T$, and there is a path from the root to each leaf in $T$, as shown in Fig.~\ref{fig:dataset_hierarchy}.
The classification of an instance belonging to $c_i \in C$ can be decomposed into classifications by multiple classifiers from the root to the leaf node corresponding to $c_i$ of the tree $T$, i.e., from coarse to fine.
An instance is correctly classified if and only if it is classified correctly at each level.

\section{Benchmark}\label{sec:benchmark}
Data are the basis of any machine learning or deep learning based task. To advance the research of Hi-FSOD, we construct the first Hi-FSOD benchmark dataset HiFSOD-Bird.

\subsection{Data Construction}
First, we collect publicly available bird images from various non-copyright websites.
Then, we perform a standard cleaning procedure to remove all low-quality images (including extremely vague ones) and duplicate images.
To acquire high-quality annotations, we recruit several well-trained annotators and ask them to annotate all the objects with bounding boxes and labels.
All the annotators have undergone professional training before annotating.
In order to make the annotation accurate, each annotator is only responsible for a specific part of data under a certain order in a period of time.
Following that, we carry out cross-checking to guarantee the annotation quality. The images that are not in our pre-defined category space are discarded.
Finally, since we are to perform 10-shot experiments, if there are less than 11 images in a category, we remove this category and all its images.
The final annotated dataset contains 176,350 high-quality images. All the bounding boxes belong to 1,432 classes of birds, falling into 32 orders, 132 families, 572 genera and 1,432 species.
Moreover, in order to facilitate the zero-shot object detection task in the future, we also acquire the textual description of each class of birds from Wikipedia, which will be released with our dataset. Fig.~\ref{fig:dataset_hierarchy} shows the taxonomy of our dataset, and Tab.~\ref{tab:dataset_summary} presents the major statistics of our dataset.

\subsection{Base/Novel Split}
\begin{figure}[t]
	\centering
	\includegraphics[width=0.9\linewidth]{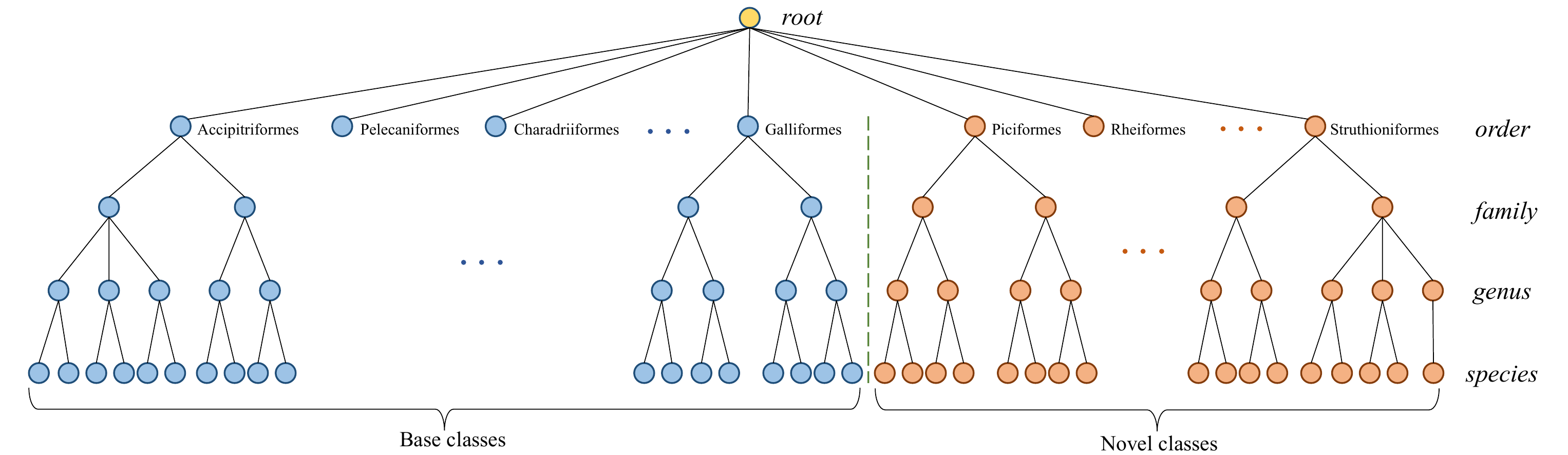}
	\caption{The hierarchical taxonomy and base/novel split in the HiFSOD-Bird dataset. We divide the base and novel classes at the order level.
	}
	\label{fig:dataset_hierarchy}
\end{figure}

\begin{figure}[t]
	\centering
	\includegraphics[width=0.9\linewidth]{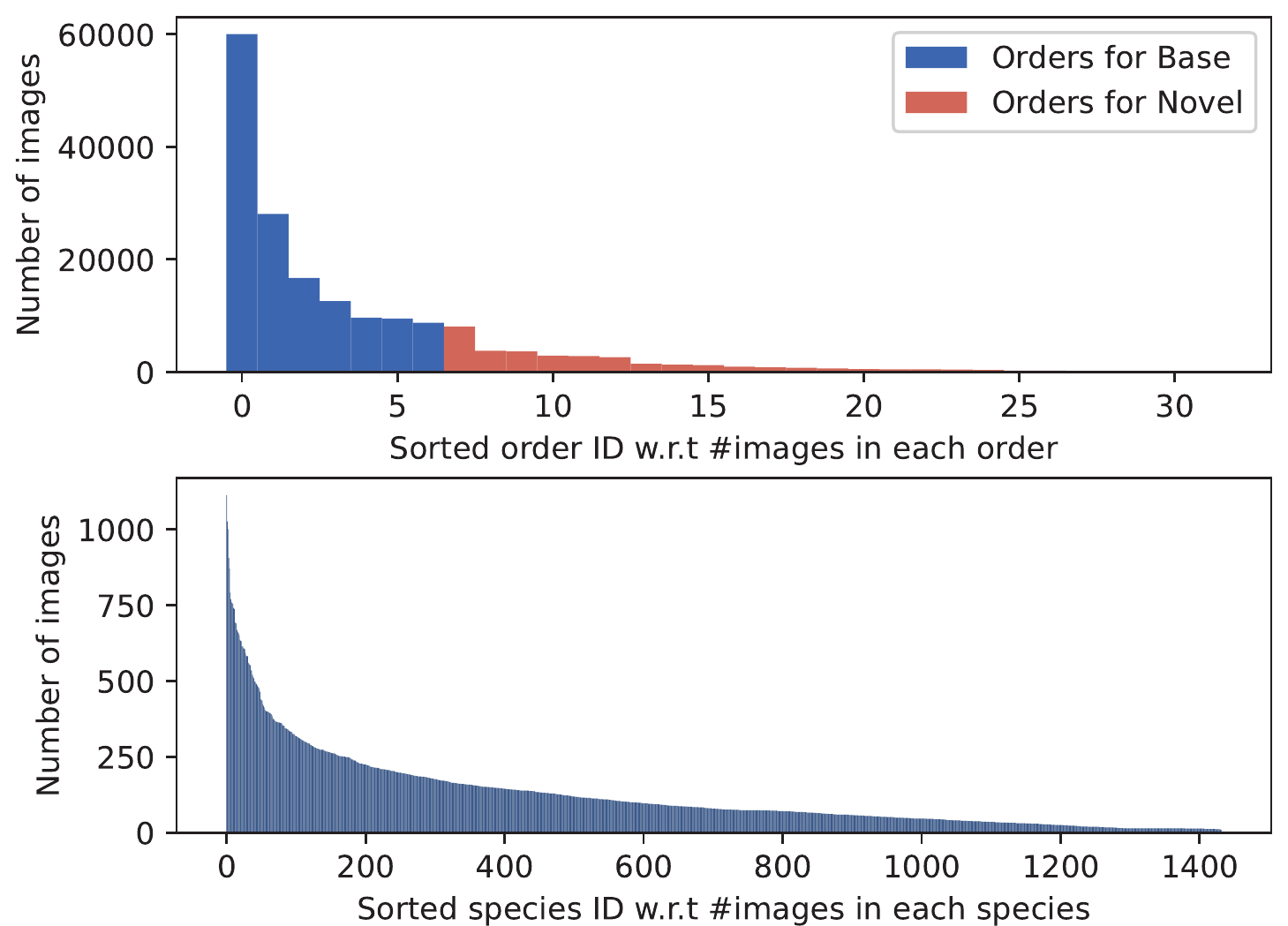}
	\caption{Distributions of images in different orders and species~(classes). The number of images within an order or class exhibits a long-tailed distribution. The classes in orders located at the distribution tail are used as novel classes.
	}
	\label{fig:dataset_dist}
\end{figure}

To perform FSOD, we need to divide all the classes into base classes and novel classes.
Since ``order'' is the top-1 level of the bird taxonomy, and birds of different orders are quite different, we divide the bird images into the base and novel classes at the order level, as shown in Fig.~\ref{fig:dataset_hierarchy}.
Moreover, the numbers of images in orders and species all exhibit a long-tail distribution as shown in Fig.~\ref{fig:dataset_dist}, which reflects the true distribution of birds in the wild.
Therefore, we sort the orders according to the number of images contained in each order and take the tail part as novel classes.
Specifically, we take the species in the top-7 orders with the largest number of images as base classes, and the species in the other 25 orders as novel classes.
Although the base classes contain birds from a less number of orders,
their images account for about 80\% of the total images.
Our base/novel division has the following advantages:
(1) The difference between base and novel classes is relatively large, since the birds of different orders are obviously different in appearance.
(2) The novel classes contain birds from a variety of different orders,
which means that on average the difference between two novel classes is also relatively large.
This is beneficial to FSOD.
(3) The species located at the distribution tail are mostly rare species that are difficult to acquire samples, which implies the potential application of our method to the monitoring and protection of endangered birds.

With the split above, we sample the images for test in each category (species). In order to perform 10-shot experiments, we sample $min$(6, $k_{c_i}$-10) images for each category as test images, where $k_{c_i}$ is the number of images in category $c_i$. Finally, a total of 8211 images are used as the test set.
All annotation files are in the COCO format for the convenience of usage.

\begin{table}[t]
	\caption{Statistics of the HiFSOD-Bird dataset. }
	\label{tab:dataset_summary}
	\resizebox{0.9\linewidth}{!}{
		\begin{tabular}{llll}
			\toprule
			& Total & Base & Novel \\
			\midrule
			\#classes (\#cls)   & 1432 & 1145 &  287 \\
			\#orders   & 32   & 7    &  25 \\
			\#families & 132  & 94   &  38 \\
			\#genera    & 572  & 436  &  136 \\
			\#species  & 1432 & 1145 &  287 \\
			\#species/\#orders & 44.75 & 163.57 & 11.48 \\
			\#images~(\#img)   & 176350 & 141239 & 35111 \\
			Range of \#img/\#cls   & [11, 1112] & [13, 1112] & [11, 1000]  \\
			Avg of \#img/\#cls    & 123.08 & 123.35 & 122.01 \\
			\#boxes             & 179042 & 143608 & 35434 \\
			box-size range              & [4, 9455160] & [4, 9455160]  & [1636, 4474202] \\
			box W/H range         & [0.274, 7]   & [0.274, 7] & [0.292, 5.27] \\
			
			\bottomrule
		\end{tabular}
	}
\end{table}

\subsection{Statistics and Characteristics}

\textbf{Statistics.}
As presented in Tab.~\ref{tab:dataset_summary}, data in HiFSOD-Bird are organized in a hierarchical structure, with a total of four levels.
HiFSOD-Bird contains 1,432 species of birds, of which 1,145 species fall into base classes and the remaining 287 species belong to novel classes. It covers more than 90\% of the world's water birds and part of the forest birds.
The number of images or species in base classes vs. that in novel classes is roughly 4:1.
Meanwhile, the size of object bounding boxes varies greatly, which makes it be challenging to detect the birds.

\textbf{Characteristics.}
(a) \emph{Hierarchical and fine-grained class space.} All categories in HiFSOD-Bird are organized according to the hierarchical taxonomy.
Meanwhile, as the level becomes deeper, the difference between categories gradually decreases and the classification difficulty increases.
(b) \emph{Long-tailed distribution}. As shown in Fig.\ref{fig:dataset_dist}, the number of images in each species  indicates a long-tailed distribution, which is consistent with the nature. Some kinds of birds are common, and their samples are easy to obtain, while most birds are less common, their samples are not easy to obtain, and there are some rare and endangered birds, whose samples are difficult to acquire.

\section{Method}\label{sec:method}

\begin{figure*}[t]
	\centering
	\includegraphics[width=0.8\linewidth]{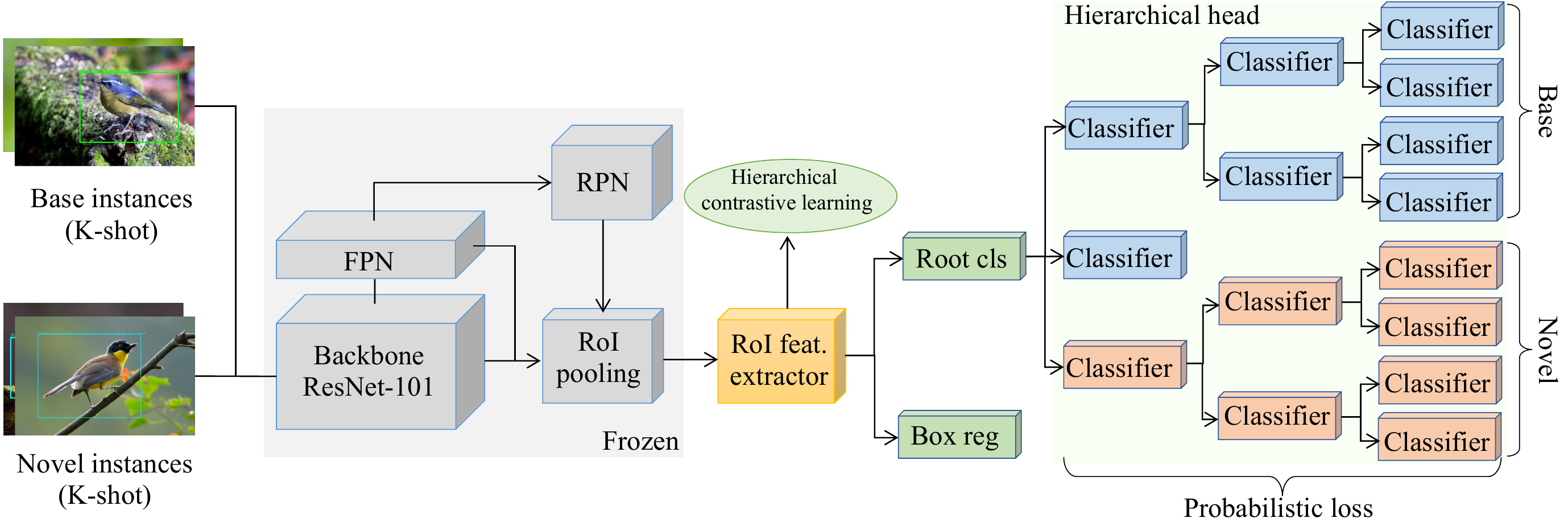}
	\caption{The framework of our proposed hierarchical few-shot object detection method HiCLPL. To make the distribution of  object  features consistent with the hierarchical taxonomy, we propose hierarchical contrastive learning. Meanwhile, to correct the errors of internal classifiers not at the lowest level in the hierarchical class tree, we design a probabilistic loss function.
	}
	\label{fig:framework}
\end{figure*}

\subsection{Overview}
Fig.~\ref{fig:framework} illustrates the framework of our method HiCLPL, where
we adopt a modified Faster R-CNN~\cite{FasterRCNN-NIPS} detection model by replacing the original classification head with a hierarchical head, where the structure of classifiers is the same as the hierarchical taxonomy $T$, whose levels are indexed from $0$~(root) to $L$~(leaf).
Each classifier $\mathcal{F}_{c_i}^j$ corresponds to a non-leaf node $\widehat{c}_i^j$ in $T$, where $j$ is the level index and $j \in [0,L-1]$, indicating that it is not needed to build classifiers for leaf nodes.
$\mathcal{F}_{c_i}^j$ needs only to discriminate the fine-grained child classes of $\widehat{c}_i^j$.
$\mathcal{F}_{c_i}^j$ is composed of two fully connected layers, and its output dimension is the number of children of  $\widehat{c}_i^j$.
Meanwhile,
we separate the regression head from the classification head, and set the regression parameters for each child class of the root node.

HiCLPL employs a simple two-stage training process.
In the first stage, we train the hierarchical Faster R-CNN with abundant base-class data (i.e., $D_{train} = D_{base}$) and the hierarchical head is built according to $T^{base}$, which is the hierarchical taxonomy of the base classes.
Then, the base detector is transferred to the novel classes through fine-tuning on a balanced dataset~\cite{TFA}, where the novel instances and sampled base instances are used as training set (i.e., $D_{train} = D_{base} \cup D_{novel} $) and the hierarchical head is built with $T^{all}$ = $T^{base} \cup T^{novel}$.

We freeze the backbone and RPN in the second stage, while unfreezing the RoI feature extractor to perform the subsequent feature distribution transformations.
In order to learn a better feature space and enhance the generalization power of the model, we propose hierarchical contrastive learning to constrain the feature space such that the feature distribution of objects is consistent with the hierarchical taxonomy.
Meanwhile, note that in the process of hierarchical classification, once an instance is classified incorrectly at a class (or node), the subsequent classifications of the instance at the descendant classes (or nodes) are meaningless. Therefore, we design a probabilistic loss to handle this problem.
Both hierarchical contrastive learning and probabilistic loss are applied in the $1^{th}$ and $2^{nd}$ stages.
Finally, we jointly optimize the hierarchical contrastive loss, the probabilistic loss and the original RPN classification and regression losses in a multi-task fashion.

\subsection{Hierarchical Contrastive Learning}
As mentioned above, in order to enhance the model's generalization power, we use hierarchical contrastive learning to constrain the feature space such that the distribution of object features is aligned with the hierarchical taxonomy $T$.
As shown in Fig.~\ref{fig:intro}(b), if an internal category $a$ has three children $a.1$, $a.2$ and $a.3$, the features of the three child categories should be close to each other in the feature space as much as possible.
We utilize a hierarchical contrastive loss to control the distribution of features, memories to hold class representations at all levels, and exponential moving average (EMA) to update memory parameters.


\subsubsection{Hierarchical Contrastive Loss}
For a box feature $x_i$ with  label $c_i$,
we denote the corresponding ground-truth path of $c_i$ in $T$ as $P_i$ 
and nodes in  $P_i$ as $\widehat{c}_i^j$, where $j$ is the level index.
For each node $\widehat{c}_i^j \in P_i$, we use a memory $\mathcal{M}_{c_i}^j$ to hold its prototype, where $j \in [0,L]$, indicating that each node on the path from root to leaf has a prototype.
Then, our goal is to maximize the agreement between box feature $x_i \in c_i$ and $\{\mathcal{M}_{c_i}^j | j \in [0,L]\}$,
and promote the deviation of $x_i$ from all the other memories at all levels.
In this way, we can not only cluster together the fine-grained classes under the same parent node,
but also make the internal categories at multiple levels be clustered according to the category hierarchy.

Inspired by supervised contrastive loss~\cite{Contrastive_Learning}, our hierarchical contrastive loss (HiCL) is defined to adapt to the hierarchical taxonomy.
Specifically, for a mini-batch containing $N$ foreground proposals with labels $\{x_i, c_i\}_{i=1}^{N}$, where $x_i \in \mathbb{R}^{2048}$ is the $i^{th}$ box feature from the RoI feature extractor without the last ReLU activation,
$c_i$ is the ground-truth class of $x_i$.
HiCL is formulated as
\begin{equation}
\mathcal{L}_{HiCL} = \frac{1}{N} \sum^{N}_{i=1} \mathcal{L}_{H_i}(x_i,c_i)
\end{equation}
\begin{small}
\begin{equation}
	\label{Eq:item_HiCL}
	\mathcal{L}_{H_i}(x_i,c_i) = \frac{-1}{  {\textstyle \sum_{j=0}^{L}}\mathcal{G}(j) }
	\sum^{L}_{j=0}\mathcal{G}(j) \log \frac{\exp ( \overline{x_i} \cdot \mathcal{M}_{c_i}^{j} / \tau ) }
	{{\textstyle \sum_{c_{\hat{i}}  \in C, \hat{j} \in [0,L] } } \exp( \overline{x_{\hat{i}}} \cdot \mathcal{M}_{c_{\hat{i}}}^{\hat{j}} / \tau ) }
	\end{equation}
\end{small}
where $\tau$ is the hyper-parameter that controls temperature,
$L$ is the height of $T$ and $\overline{x_i}=\frac{x_i}{\left \| x_i \right \|}$ is the normalized box feature.
$\mathcal{G}(\cdot)$ is the function to regularize the strength of aggregation at different levels. We found that it is better to increase the strength of aggregation with the increase of the level $j$, and we set $\mathcal{G}(j) = j$ in our experiments. We present the experimental results on how the function impacts the performance in Sec.~\ref{Sec:AB_G} .

\begin{figure}[t]
	\centering
	\includegraphics[width=0.9\linewidth]{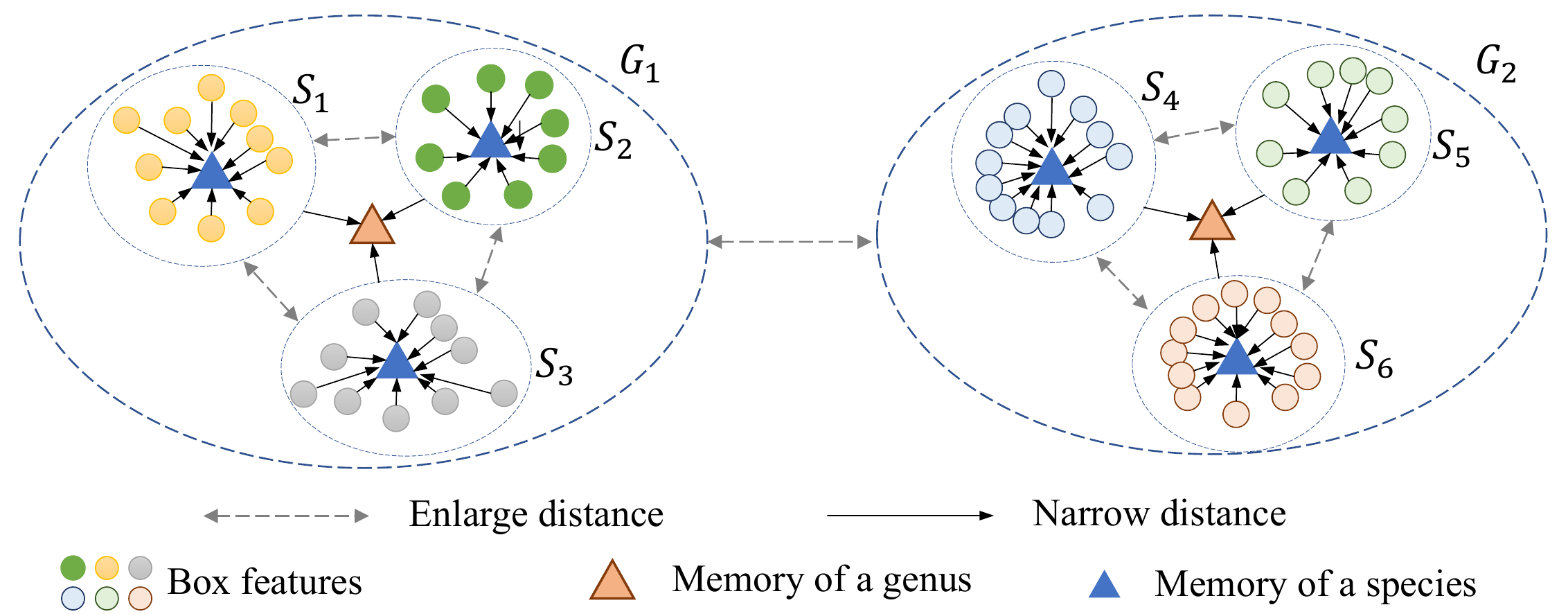}
	\caption{Illustration of hierarchical contrastive learning. We constrain the feature distribution of objects in  feature space to be consistent with the hierarchical taxonomy.
	}
	\label{fig:Hierarchical_Contrastive}
\end{figure}

In Eq.~(\ref{Eq:item_HiCL}), we try to narrow the distance between $x_i$ and the memories it belongs to from the leaf memory $\mathcal{M}_{c_i}^{L}$ to the root memory $\mathcal{M}_{c_i}^{0}$ with different aggregation strength $\mathcal{G}(\cdot)$, and enlarge the distances to the other memories. In this way, the object features will be aggregated according to the hierarchical taxonomy $T$ and the distribution of the feature space is consistent with the hierarchy of the categories. Fig.~\ref{fig:tSNE} shows the effectiveness  by t-SNE visualization.

\subsubsection{Update of Memory Parameters}
The memories described above store the latest representation of each node in $T$. Hence, the  memories need to be updated during training.
Here, we use exponential moving average (EMA)~\cite{EMA,Episodic_Memory,Memory_REID} to update the parameters of memories.
Concretely, for a box feature $x_i$ with ground-truth class $c_i$ and ground-true path $P_i$ in $T$, each memory $\mathcal{M}_{c_i}^j, j \in [0,L]$ in path $P_i$ will be updated as follows:
\begin{equation}
\label{Eq:EMA_update}
\mathcal{M}_{c_i}^j \gets f(j) \mathcal{M}_{c_i}^j + [1-f(j)] \overline{x_i}, \quad j \in [0,L]
\end{equation}
\begin{equation}
	\label{Eq:update_rate}
	f(j) = 1 - {\epsilon}^{L - j + 1}
\end{equation}
where $\overline{x_i}$ is the normalized $x_i$ as that in Eq.~(\ref{Eq:item_HiCL}).
$f(j)$ is the momentum coefficient for updating memories and
$\epsilon$ is a hyper-parameter to control the value of $f(j)$, which empirically takes a small value ($<1.0$) such as 0.1.
Eq.~(\ref{Eq:EMA_update}) indicates that for a box feature $x_i$, we update all the memories in a bottom up way along its ground-truth path in $T$.
In other words, the memory of a node aggregates the box features of all its descendant nodes.
Meanwhile, we can see that the momentum coefficients $f(j)$ of the memories at different levels are different, as the memories at different levels are updated by using different numbers of their most recent box features. An upper-level memory needs much more recent features to be updated than any of its descendant memories, due to covering more descendant categories.


\subsection{Probabilistic Loss}
A common problem in hierarchical classification is that if a node classifies incorrectly, the classifications of its descendant nodes will be meaningless.
Therefore, we design a mechanism to enable a node to judge and correct the classifications of its parent node.

For a box feature $x_i$ with ground-true path $P_i$ in $T$, $\{\widehat{c}_i^j | \widehat{c}_i^j \in P_i, j \in [0,L-1] \}$  are internal nodes in $P_i$.
As mentioned before, each internal node corresponds to a classifier. For an arbitrary classifier $\mathcal{F}_{c_k}^j$,
our aim is to reduce the outputs for all classes of $\mathcal{F}_{c_k}^j$ if the ground-truth path of $x_i$ does not pass through $\mathcal{F}_{c_k}^j$. Then, the product of probabilities on any wrong path will be restrained to less than the product of probabilities on the correct path, so the classification error of the parent node can be corrected.

\begin{figure}[t]
	\centering
	\includegraphics[width=1\linewidth]{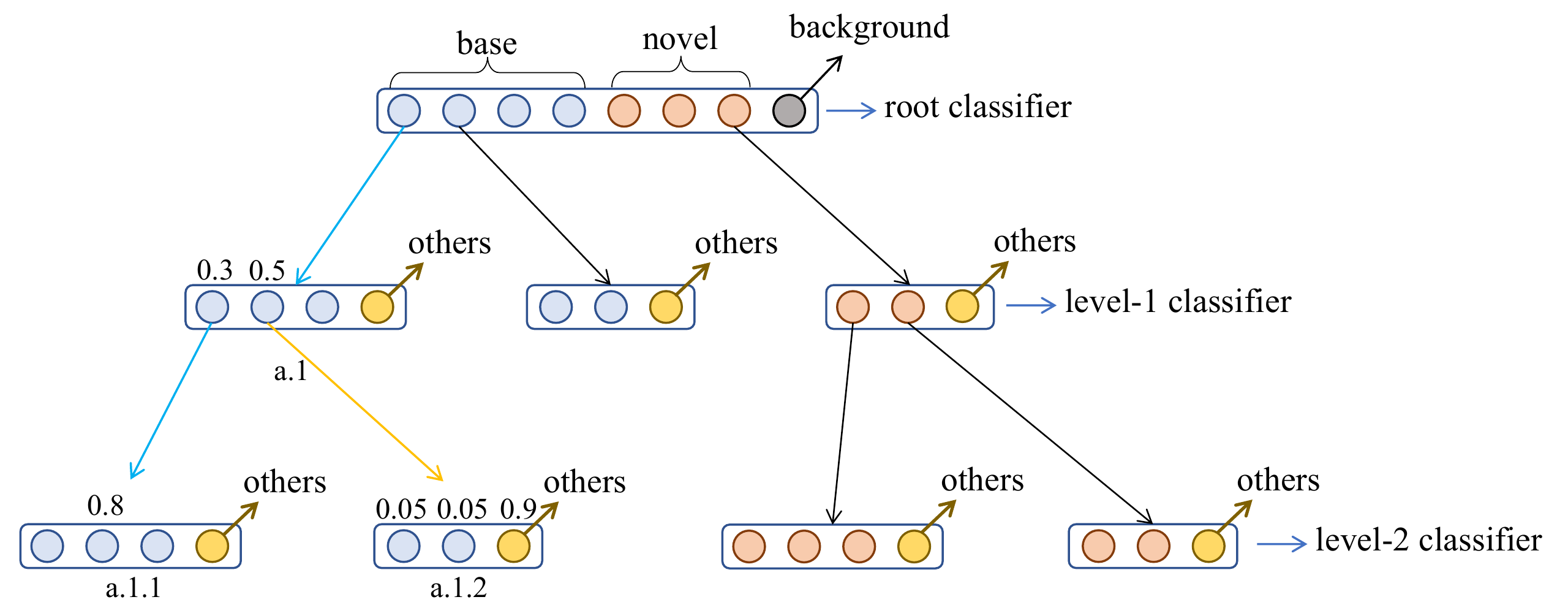}
	\caption{A 3-level structure of classifiers. We add a virtual class ``others'' for each classifier to make the classifier classify a sample not belonging to its predefined classes into the ``others'' class, so as to correct the errors of upper nodes. The blue arrow is the ground-truth path. The yellow arrow is an error of node $a.1$, which will be corrected by node $a.1.2$ via classifying the sample to its ``others'' class.
	}
	\label{fig:classifier_structure}
\end{figure}

To achieve this, we add an ``others'' virtual class for each classifier.
The structure of classifiers is shown in Fig.~\ref{fig:classifier_structure}.
If a sample does not belong to any predefined class of the classifier, it will be classified to the ``others'', such as the classifier $a.1.2$ does in Fig.~\ref{fig:classifier_structure}.
When inferring a model, we do not only go along a path of the largest score when classifying from root to leaf, but also use beam search~\cite{Beam-Search}, taking multiple paths into consideration so that the potentially correct path can be discovered and nodes with classification errors can be abandoned later because of their children's low scores.

To train the classifier $\mathcal{F}_{c_i}^j$ with the virtual class ``others'', we need to sample instances that do not belong to $\mathcal{F}_{c_i}^j$'s predefined classes (labeled from 0 to $d_i^j$-1) for the ``others'' class and assign them the label $d_i^j$, where $d_i^j$ is the number of children for node $\widehat{c}_i^j$.
For $\mathcal{F}_{c_i}^j$, if all instances not belonging to its original classes are assigned to ``others'', then extreme class imbalance will happen. To solve this problem, we introduce the \emph{probabilistic loss} (PL).

Formally,
for a mini-batch containing $N$ foreground proposals with labels  $\{x_i,c_i,P_i\}_{i=1}^N$, where $x_i$ is the box feature, $c_i$ is the ground truth label and $P_i$ is the ground-truth path, the probabilistic loss is formulated as

\begin{equation}
	\label{Eq:loss_prob}
	\begin{split}
		\mathcal{L}_{prob} = \frac{1}{MN} \sum_{c_k \in C, j \in [0,L-1]}
		 \sum_{i=1}^{N}
		[
		\mathbb{I}(\widehat{c}_k^j \in P_i) \cdot \mathcal{L}_{CE}(c_k, j,  i)  \\
		+ \mathbb{I}(\widehat{c}_k^j \notin P_i) \cdot \mathbb{I}(\mathcal{P}_k^j) \cdot \mathcal{L}_{oth}(c_k, j, i)
		]
	\end{split}
\end{equation}

\begin{equation}
	\label{Eq:CE_loss}
	\mathcal{L}_{CE}(c_k, j, i) = - y_i^{k,j} \cdot \log(\mathcal{F}_{c_k}^j(x_i))
\end{equation}

\begin{equation}
	\mathcal{L}_{oth}(c_k, j, i) = - d_k^j \cdot \log(\mathcal{F}_{c_k}^j(x_i))
\end{equation}

Eq.~(\ref{Eq:loss_prob}) indicates that we train all classifiers sequentially. If $x_i$ belongs to the internal node $\widehat{c}_k^j$, i.e., $\widehat{c}_k^j \in P_i$, we train $\mathcal{F}_{c_k}^j$ with cross entropy loss as in Eq.~(\ref{Eq:CE_loss}), where $y_i^{k,j}$ is the corresponding label of $x_i$ for $\mathcal{F}_{c_k}^j$. Otherwise, we train $\mathcal{F}_{c_k}^j$ to classify $x_i$ into ``others'' with a probability $\mathcal{P}_k^j$.
Here, $\mathbb{I}(\mathcal{P}_k^j)$ is the indicator function that takes true with a probability $\mathcal{P}_k^j$. $d_k^j$ is the number of children of node $\widehat{c}_k^j$, which is also the label (or index) of the ``others'' class. $M$ is the number of classifiers, each of which is trained once in an iteration.

$\mathcal{P}_k^j$ reflects the distribution of categories.
We assume the total number of instances in the dataset is $\mathcal{N}$,
the ground-truth path of instance $x_i$ is $P_i$.
$\mathcal{P}_k^j$ is calculated as follows:
\begin{equation}
	\label{Eq:prob_cal}
	\mathcal{P}_k^j = \frac{ \sum_{i=1}^{\mathcal{N} } \mathbb{I}( \widehat{c}_k^j \in P_i ) }{\mathcal{N} } \cdot \beta
\end{equation}

In Eq.~(\ref{Eq:prob_cal}), we calculate the proportion of instances contained in each internal node, and then multiply the proportion by an adjustment factor $\beta$, which is used as the probability to train the ``others'' class. In this way, the probability that the classifier trains the ``others'' class is related to the number of samples it contains, which can effectively alleviate the class imbalance problem.

In order to make our correction mechanism work better, we need to consider multiple child nodes of a parent node during model inference.
The prediction result of box feature $x_i$ is
\begin{equation}
	\widehat{y}_i = argmax\{\prod_{j=1}^{L} p_i^{j,k} | p_i^{j,k} \in P_k, P_k \in \mathbf{P} \}
\end{equation}
where $\mathbf{P}$ is the set of all paths in $T$, $P_k$ is the $k^{th}$ path in $\mathbf{P}$ and $p_i^{j,k}$ is the prediction score at the $j^{th}$ level of path $P_k$ for $x_i$ after \textit{Softmax}.

However, it is time consuming to recursively run the procedure above.
Here, we utilize beam search~\cite{Beam-Search} to balance time consumption and performance. 
Finally, standard threshold screening and non-maximum suppression~(NMS)
are applied to getting outputs. 

\begin{table*}[t]
	\caption{Performance of novel classes on HiFSOD-Bird dataset.}
	\label{tab:results_comp}
	\resizebox{1\linewidth}{!}{
		\begin{tabular}{c|ccc|ccc|ccc|ccc|ccc}
			\toprule
			\multirow{2}{*}{Method / Shot} &
			\multicolumn{3}{c|}{1-shot} &
			\multicolumn{3}{c|}{2-shot} &
			\multicolumn{3}{c|}{3-shot} &
			\multicolumn{3}{c|}{5-shot} &
			\multicolumn{3}{c}{10-shot}
			\\
			&  $AP$ & $AP_{50}$ & $AP_{75}$ &  $AP$ & $AP_{50}$ & $AP_{75}$ &  $AP$ & $AP_{50}$ & $AP_{75}$ &  $AP$ & $AP_{50}$ & $AP_{75}$ &  $AP$ & $AP_{50}$ & $AP_{75}$ \\
			\midrule
			Attention-RPN~\cite{Attention-RPN} &12.81&18.55&14.89& 19.10& 26.31&23.24& 15.28&21.22&18.54&25.20& 34.40& 30.71&28.74&39.25& 35.08 \\
			TFA~\cite{TFA} & 5.16 & 7.41 & 5.83 & 24.47&32.70&29.51& 29.84&39.76&35.91&37.16&49.61&44.31& 42.74&56.62&50.30 \\
			
			GFSOD~\cite{GFSOD} &  10.58 &22.88&7.98&13.61&29.94&9.32&14.70&33.95&10.36&15.15&37.20&9.32&20.66&47.95&13.42\\
			
			FSCE~\cite{FSCE} &  11.83&15.78&12.35& 26.89&35.31&32.13&30.47&41.10&36.15& 38.20&51.28&45.94&42.90&56.79&50.88 \\
			
			HiCLPL~(Ours) &  \textbf{18.95} &\textbf{26.09} &\textbf{22.31} &
			\textbf{28.50} &\textbf{37.83}&\textbf{34.17}&
			\textbf{31.77} & \textbf{42.95} &\textbf{37.82} &
			\textbf{39.37} & \textbf{52.93} & \textbf{47.31} &
			\textbf{43.54} & \textbf{57.74}  & \textbf{51.57}
			\\
			\bottomrule
		\end{tabular}
	}
\end{table*}

\begin{table}[t]
	\caption{Performance for base and novel classes. 
	}
	\label{tab:results_comp_base_novel}
	\resizebox{0.8\linewidth}{!}{
		\begin{tabular}{cc|cc}
			\toprule
			Shot & Method & Base AP50 & Novel AP50
			\\
			\midrule
			\multirow{4}{*}{1} & TFA~\cite{TFA} & 72.79 & 7.41 \\
			& GFSOD~\cite{GFSOD} & 71.47 & 22.88 \\
			& FSCE~\cite{FSCE} & 72.72   & 15.78 \\
			& HiCLPL~(Ours)     & \textbf{75.72}   & \textbf{26.09}  \\
			\bottomrule
			\midrule
			\multirow{4}{*}{5} & TFA~\cite{TFA} & 74.54 & 49.61  \\
			& GFSOD~\cite{GFSOD} & 71.93 & 37.20 \\
			& FSCE~\cite{FSCE} & 74.16   & 51.28 \\
			& HiCLPL~(Ours)     & \textbf{75.97}   & \textbf{52.93}  \\
			\bottomrule
			
		\end{tabular}
	}
\end{table}

\subsection{Training Loss}
The total loss of our model used for training is as follows:
\begin{equation}
	\mathcal{L} = \mathcal{L}_{RPN_{Cls}} + \mathcal{L}_{RPN_{Reg}} + \mathcal{L}_{Reg}
	+ \lambda_1 \mathcal{L}_{HiCL} + \lambda_2 \mathcal{L}_{prob}
\end{equation}
where the first two terms are the cross entropy loss and regression loss of RPN respectively. $\mathcal{L}_{Reg}$ is the loss for box regression in Fig.~\ref{fig:framework}. $\mathcal{L}_{HiCL}$ is our hierarchical contrastive loss. $\mathcal{L}_{prob}$ is the probabilistic loss. We set $\lambda_1 = 0.5$ and $\lambda_2 = 1$ in our experiments to balance different loss functions.

\section{Experiments}\label{sec:experiments}
\subsection{Implementation Details}
We use Faster-RCNN~\cite{FasterRCNN-NIPS} with ResNet-101~\cite{Resnet} and Feature Pyramid Network~\cite{FPN} as backbone. All models are trained on 4 NVIDIA RTX 3090 in parallel with batch size 16. We employ SGD with momentum 0.9 as the optimizer.
For different shots, we take a different number of iterations.
The height of taxonomy $L$ is set to 4. The hyper-parameter $\tau$ in Eq.~(\ref{Eq:item_HiCL}) is set to 0.2, $\epsilon$ in Eq.~(\ref{Eq:update_rate}) is set to 0.1 and $\beta$ in Eq.~(\ref{Eq:prob_cal}) is set to 0.5.
Code and dataset are presented in \href{Github}{https://github.com/zhanglu-cst/HIFSOD}.


\subsection{Comparison with Existing Methods}
We select several typical FSOD algorithms as baselines for comparison and make minor modifications to them for fitting our HiFSOD-Bird dataset, including Attention-RPN~\cite{Attention-RPN}, TFA~\cite{TFA}, GFSOD~\cite{GFSOD} and FSCE~\cite{FSCE}.
Results of novel classes on HiFSOD-Bird are shown in Tab.~\ref{tab:results_comp}.
We strictly follow a consistent evaluation protocol~\cite{TFA} of COCO to ensure fair comparison.
Metrics of $AP$, $AP_{50}$ and $AP_{75}$ are used.
As we can see, our method HiCLPL outperforms all the compared existing methods in any shot and on all metrics, demonstrating the effectiveness and superiority of our method in
hierarchical few-shot object detection.
Especially, when the number of shots is extremely low, our method outperforms the baselines a large margin. Concretely,
for 1-shot, our method surpasses Attention-RPN~\cite{Attention-RPN} by 6.14, 7.54 and 7.42 points in terms of $AP$, $AP_{50}$ and $AP_{75}$ respectively.
As the shot number grows, our method still keep advantageous.
For 5-shot, our method outperforms FSCE~\cite{FSCE} by 1.17, 1.65 and 1.37 points in $AP$, $AP_{50}$ and $AP_{75}$ respectively.


Following previous works~\cite{TFA,FSCE}, we also report the performance on base classes for all methods. Results are shown in Tab.~\ref{tab:results_comp_base_novel}.
We can see that our method also exceeds all the baselines on the base classes, proving that our method is also effective in detecting the base classes.
Due to the large number of base classes in HiFSOD-Bird, Attention-RPN~\cite{Attention-RPN} does not work due to out of memory, so its base-class results are not reported here.

\subsection{Ablation Study}
We check the effects of various components.
Unless otherwise specified, the ablation studies are carried out in 2-shot setting.

\begin{table}[t]
	\caption{Effects of different components. HiHead: Hierarchical Head. HiCL: Hierarchical Contrastive Learning. Prob Loss: Probabilistic Loss. }
	\label{tab:ab_all}
	\resizebox{0.85\linewidth}{!}{
		\begin{tabular}{ccc|cc}
			\toprule
			HiHead &  HiCL &  Prob Loss &  Base AP50  & Novel AP50 \\
			\midrule
			&   &     &  73.52  &   32.70 \\
			$\checkmark$  &    &    &  72.05  & 30.81 \\
			$\checkmark$  &    &  $\checkmark$  &  73.24  & 32.53 \\
			$\checkmark$  &  $\checkmark$  &  &  75.79  & 36.16 \\
			$\checkmark$  &  $\checkmark$  &  $\checkmark$  &  \textbf{76.12}  & \textbf{37.83} \\
			\bottomrule
		\end{tabular}
	}
\end{table}

\subsubsection{\textbf{Effects of different components}}
We investigate the effects of different modules and summarize the results in Tab.~\ref{tab:ab_all}.
The implementation details are:
(1) The baseline without hierarchical head (the first line) is TFA~\cite{TFA}.
(2) For models without hierarchical contrastive learning, we remove the supervision from RoI feature extractor. The RoI feature extractor is supervised by subsequent classifiers and regressors.
(3) For models without probabilistic loss, we use standard cross-entropy loss to train the classifiers.

From Tab.\ref{tab:ab_all}, we can see that
after adding hierarchical head, the performance is worsened, which is due to the overfitting caused by the excessive number of parameters in the hierarchical head. However,
after HiCL is introduced,
the performance is greatly improved.
Compared with baseline with only hierarchical head, the model with HiCL boosts the performance on base classes from 72.07 to 75.79 and that on novel classes from 30.81 to 36.16, demonstrating the effectiveness of
hierarchical contrastive learning.
The probabilistic loss can also boost the performance to a certain extent, concretely, bringing about one point improvement on both base classes and novel classes.

\subsubsection{\textbf{Ablation study on function $\mathcal{G}(\cdot)$}}
\label{Sec:AB_G}
Function $\mathcal{G}(\cdot)$ is used to adjust the aggregation strength of different levels.
We conduct experiments with different forms of $\mathcal{G}(\cdot)$,
the results are presented in Tab.~\ref{tab:ab_func_g}.
We can see that when we apply the same aggregation strength to each level ($\mathcal{G}(x) = 1$), the performance is not good.
This is because similar strengths may cause multiple subclasses under the same parent node to be too close, making it difficult to distinguish the subclasses.
When we use $\mathcal{G}(x) = x^2$, large aggregation strengths will be applied to the subclasses, but the aggregation strength for the parent node will be reduced.

\begin{table}[t]
	\caption{Ablation study on the function $\mathcal{G}(\cdot)$.  }
	\label{tab:ab_func_g}
	\resizebox{0.8\linewidth}{!}{
		\begin{tabular}{c|cc}
			\toprule
			Adjust Function $\mathcal{G}(\cdot)$ & Base AP50 & Novel AP50 \\
			\midrule
			$\mathcal{G}(x) = 1$  & 75.47 & 37.28 \\
			$\mathcal{G}(x) = x$  & 76.12 & 37.83 \\
			$\mathcal{G}(x) = x^2$ & 75.87 & 37.60 \\
			\bottomrule
		\end{tabular}
	}
\end{table}

\subsubsection{\textbf{Ablation study on hyper-parameters in HiCL}}
\label{Sec:AB_hyper_HiCL}
Here, we check the effects of two hyper-parameters in hierarchical contrastive learning, which are the temperature $\tau$ for contrastive learning in Eq.~(\ref{Eq:item_HiCL}) and $\epsilon$ for
controlling the value of momentum coefficient $f(j)$ in Eq.~(\ref{Eq:update_rate}).
Results are given in Tab.~\ref{tab:ab_hyper_hicl}, where we can see that $\tau = 0.2$ performs better than the other settings.
As for $\epsilon$, when $\epsilon = 0.1$, the performance is best.
A large $\epsilon$ means that the memory update amplitude is large, which may cause instability in training. But
a much smaller $\epsilon$ may make prototype update too slower.

\begin{table}[h]
	\caption{Ablation on hyper-parameters in HiCL.}
	\label{tab:ab_hyper_hicl}
	\resizebox{0.75\linewidth}{!}{
		\begin{tabular}{c|cc}
			\toprule
			hyper-parameter  & Base AP50 & Novel AP50  \\
			\midrule
			$\tau = 0.05, \epsilon = 0.1$ & 75.69 & 37.65\\
			$\tau = 0.2, \epsilon = 0.1$  & 76.12 & 37.83 \\
			$\tau = 0.5, \epsilon = 0.1$  & 76.05 & 37.78\\
			\hline
			$\tau = 0.2, \epsilon = 0.01$ & 75.97 & 37.50 \\
			$\tau = 0.2, \epsilon = 0.5$  & 74.71 & 36.62\\
			\bottomrule
		\end{tabular}
	}
\end{table}

\subsubsection{\textbf{Visualization effect of HiCL}}
\begin{figure}[t]
	\centering
	\includegraphics[width=0.9\linewidth]{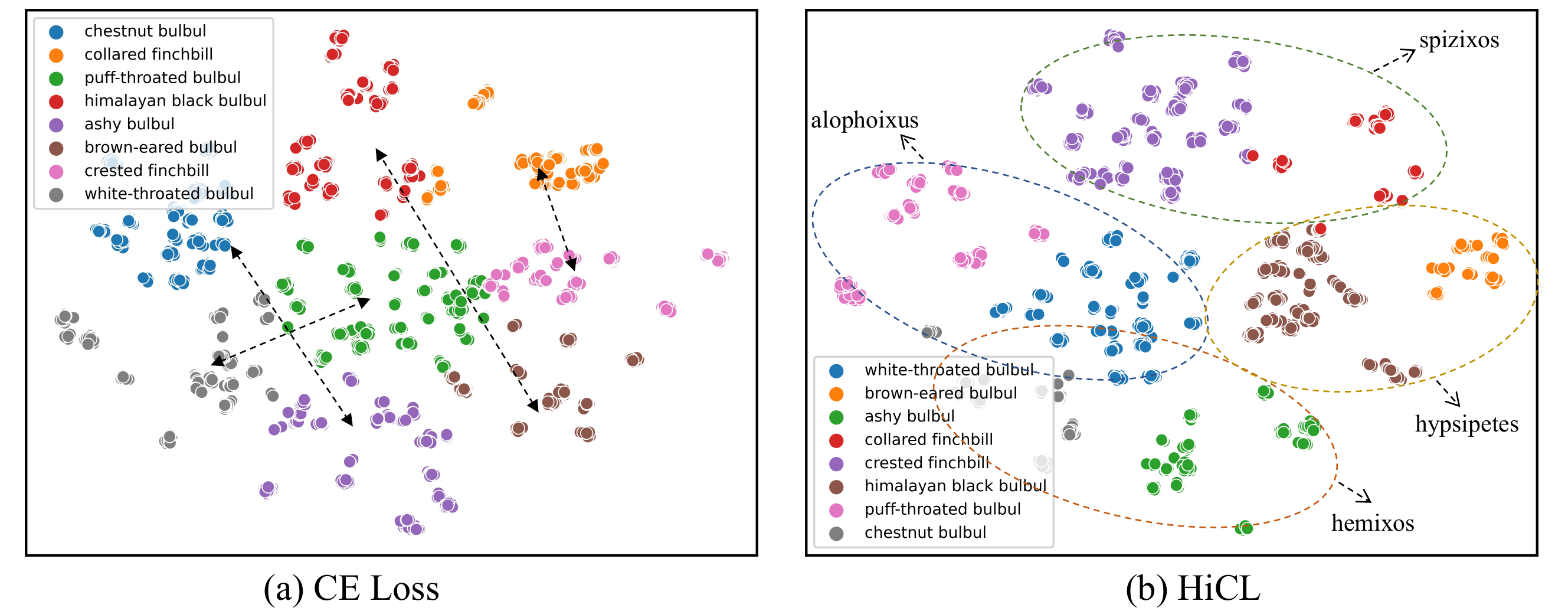}
	\caption{t-SNE visualization of box features. We use arrows to indicate different species under the same genus. (a) The features learnt by CE loss are messy. (b) Our HiCL can effectively constrain the feature space.
	}
	\label{fig:tSNE}
\end{figure}

Fig.~\ref{fig:tSNE} shows the visual results of (a) cross entropy loss and (b) our hierarchical contrastive learning.
Here, we select the species under ``Pycnonotidae'' family for visualization, which are
``ashy bulbul'' and ``chestnut bulbul'' under the ``hemixos'' genus,
``white-throated bulbul'' and ``puff-throated bulbul'' under the ``alophoixus'' genus,
``himalayan black bulbul'' and ``brown-eared bulbul'' under the ``hypsipetes'' genus,
``crested finchbill'' and ``collared finchbill'' under the ``spizixos'' genus.
In the feature space learnt by the naive cross entropy loss, the species of different genera are closely located, resulting in poor generalization.
With HiCL, the features of species are organized according to the genera they belong to.
Moreover, the feature space forms a more compact structure with more distinctive boundary, which greatly enhances the generalization power of the model.

\subsubsection{\textbf{Ablation study on probabilistic loss}}
First, we perform ablation study on the hyper-parameter $\beta$ in Eq.~(\ref{Eq:prob_cal}) of the probabilistic loss.
$\beta$ is an adjustment factor that controls the probability of training the virtual ``others'' class.
Results of different $\beta$ values are shown in Tab.~\ref{tab:ab_hyper_beta}.
We can see that the model performs best when $\beta = 0.5$.
When $\beta$ is close to 0 (e.g. $\beta = 0.1$), the probability loss is degenerating to a cross-entropy loss. As a result, performance is degraded. However,
when $\beta$ is too large (e.g. $\beta = 1.0$), the number of samples used for training the ``others'' class for each classifier will be too large, resulting in a certain degree of class imbalance.

\begin{figure}[t]
	\centering
	\includegraphics[width=1.0\linewidth]{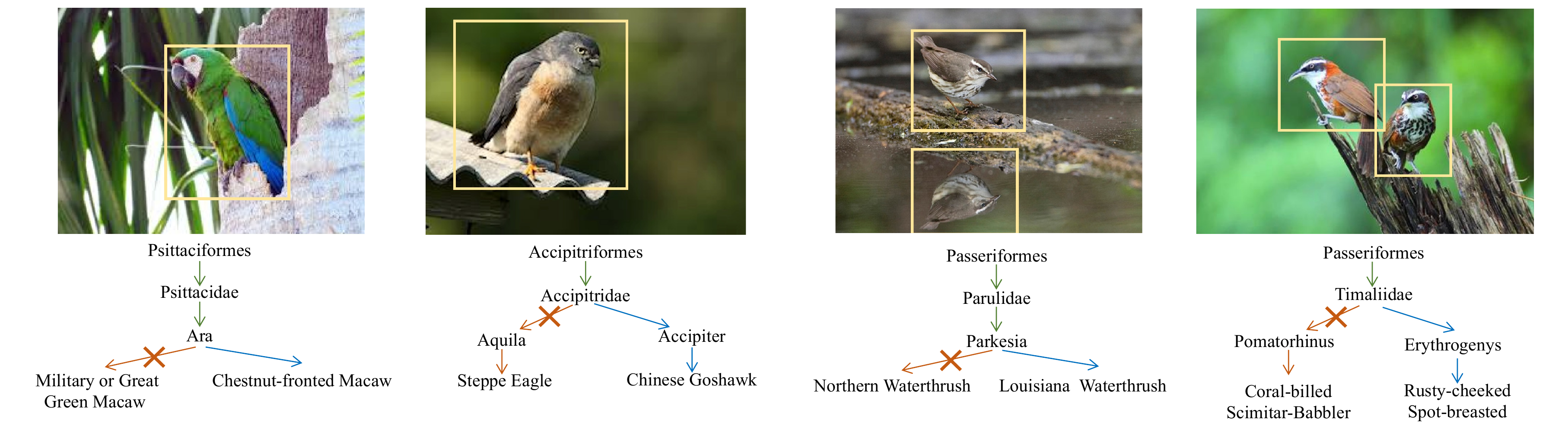}
	\caption{Some cases of misclassification corrected by our probabilistic loss. The cross indicates where the original CE loss classification goes wrong.
	}
	\label{fig:correct}
\end{figure}

Then, we show some cases of wrong classification corrected by our probabilistic loss in Fig.~\ref{fig:correct}, which verify that the probabilistic loss can recognize and correct the errors of parent nodes. Essentially, this is an embodiment of ensemble learning.

\begin{table}[t]
	\caption{Ablation on $\beta$ in the probabilistic loss.}
	\label{tab:ab_hyper_beta}
	\setlength{\tabcolsep}{1.5mm}{
		\begin{tabular}{c|cc}
			\toprule
			$\beta$  & Base AP50 & Novel AP50  \\
			\midrule
			0.1  & 75.83 & 36.20 \\
			0.5  & 76.12 & 37.83 \\
			1    & 76.07 & 37.79 \\
			\bottomrule
		\end{tabular}
	}
\end{table}

\section{Conclusion}
This paper proposes a new problem called hierarchical few-shot object detection~(Hi-FSOD), which is a nontrivial extension to the existing FSOD task.
To solve Hi-FSOD, on the one hand, we establish the first large-scale and high-quality Hi-FSOD benchmark dataset HiFSOD-Bird, which contains 176,350 wild-bird images falling to 1,432 categories that are organized into a 4-level taxonomy.
On the other hand, we develop the first Hi-FSOD method HiCLPL, where hierarchical contrastive learning is proposed to constrain the feature space and a probabilistic loss is designed to correct the errors of parent nodes.
Extensive experiments on the benchmark dataset demonstrate the effectiveness and advantage of the proposed method.

\textbf{Acknowledgement.} Yang Wang and Jihong Guan were partially supported by NSFC (No.~U1936205) and Key R\&D Projects of the Ministry of Science and Technology of China~(No.~2021YFC3300300).

%
\bibliographystyle{ACM-Reference-Format}
\bibliography{sample-base}


\end{document}